\documentclass[sigconf]{acmart}
\usepackage[ruled,vlined]{algorithm2e}
\usepackage{tabularx}
\usepackage{multirow}

\AtBeginDocument{%
  \providecommand\BibTeX{{%
    \normalfont B\kern-0.5em{\scshape i\kern-0.25em b}\kern-0.8em\TeX}}}

\copyrightyear{2020}

\setcopyright{none}
\renewcommand\footnotetextcopyrightpermission[1]{} 

\begin{document}

\title{Visual Summarization of Lecture Video Segments for Enhanced Navigation}

\author{Mohammad Rajiur Rahman}
\email{mrahman13@uh.edu}
\orcid{0002-4462-0274}
\affiliation{%
  \institution{University of Houston}
  \streetaddress{3551 Cullen Blvd.}
  \city{Houston}
  \state{TX}
  \postcode{77204-3010}
}

\author{Jaspal Subhlok}
\email{jaspal@uh.edu}
\affiliation{%
  \institution{University of Houston}
  \streetaddress{3551 Cullen Blvd.}
  \city{Houston}
  \state{TX}
  \postcode{77204-3010}
}

\author{Shishir K. Shah}
\email{sshah5@uh.edu}
\affiliation{%
  \institution{University of Houston}
  \streetaddress{3551 Cullen Blvd.}
  \city{Houston}
  \state{TX}
  \postcode{77204-3010}
}

\renewcommand{\shortauthors}{Rahman, et al.}

\begin{abstract}



Lecture videos are an increasingly important learning resource for higher education. However, the challenge of quickly finding the content of interest in a lecture video is an important limitation of this format. This paper introduces visual summarization of lecture video segments to enhance navigation. A lecture video is divided into segments based on the frame-to-frame similarity of content. The user navigates the lecture video content by viewing a single frame visual and textual summary of each segment. The paper presents a novel methodology to generate the visual summary of a lecture video segment by computing similarities between images extracted from the segment and employing a graph-based algorithm to identify the subset of most representative images. The results from this research are integrated into a real-world lecture video management portal called Videopoints. To collect ground truth for evaluation, a survey was conducted where multiple users manually provided visual summaries for 40 lecture video segments selected from a diverse set of courses. The users also stated whether any images were not selected for the summary because they were similar to other selected images. The graph based algorithm for identifying summary images achieves 78\% precision and 72\% F1-measure with frequently selected images as the ground truth, and 94\% precision and 72\% F1-measure with the union of all user selected images as the ground truth.  For 98\% of algorithm selected visual summary images, at least one user also selected that image for their summary or considered it similar to another image they selected. Over 65\% of automatically generated summaries were rated as good or very good by the users on a 4-point scale from poor to very good. Overall, the results establish that the methodology introduced in this paper produces good quality visual summaries that are practically useful for lecture video navigation.

\end{abstract}





\maketitle

\section{Introduction}






 \begin{figure*}
\includegraphics[width=5in]{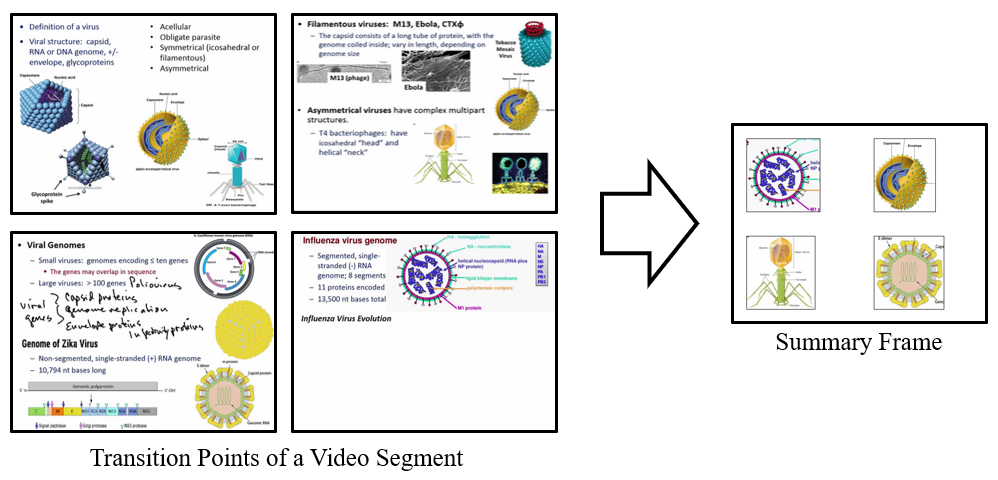}
  \caption{Visual Content Summarization}
  \Description{Summary is created from the visual content extracted from transition frames of a video segment}
  \label{fig:teaser}
\end{figure*}
Recorded lecture videos are gaining popularity as a core tool for distance learning as well as a supplementary tool for face-to-face learning. Popular commercial lecture video hosting platforms include Echo360\cite{Echo360}, Kaltura\cite{Kaltura}, and Panopto\cite{Panopto}. When classroom lectures are captured on video and made available to students, they make use of them, enjoy using them, and perceive them to be a valuable learning tool \cite{Abowd_1999,leciaBarker:studentPerceptions,Brandsteidl_Mayerhofer_Seidl_Huemer_2012,Brotherton_Abowd_2004,Chandra_2007,Defranceschi_Ronchetti_2011, Dickson_2011, Dickson_Warshow_Goebel_Roache_Adrion_2012, Johnston_Massa_Burne_2013,Lancaster_McQueeney_VanAmburgh_2011, Odhabi_Nicks-McCaleb_2011, Soong_Chan_Cheers_2006, Traphagan_Kucsera_Kishi_2010, tuna:developmentandevaluation}. Lecture videos positively impact traditional lecture based coursework as well as online learning. In situations where videos are provided prior to class, and class time is instead used for active learning, positive effects include increased student retention, learning, and engagement \cite{Carlisle_2010,Foertsch_Moses_Strikwerda_Litzkow_2002,Heilesen_2010,Lage_Platt_Treglia_2000,Sorden_Ramirez-Romero_2012}. Students value videos for allowing them to review at their own pace  \cite{Brandsteidl_Mayerhofer_Seidl_Huemer_2012, Dickson_Warshow_Goebel_Roache_Adrion_2012, Johnston_Massa_Burne_2013, Toppin_2011,Traphagan_Kucsera_Kishi_2010} and typically report that their use of videos had a positive impact on grades and overall course satisfaction \cite{Brandsteidl_Mayerhofer_Seidl_Huemer_2012,Defranceschi_Ronchetti_2011,Dickson_Warshow_Goebel_Roache_Adrion_2012,Johnston_Massa_Burne_2013,Lancaster_McQueeney_VanAmburgh_2011,OBannon_Lubke_Beard_Britt_2011,Traphagan_Kucsera_Kishi_2010} .

This research is conducted in the context of
Videopoints project at the University of Houston whose central goal is to ease navigation of lecture videos, making them a companion resource for learning, similar to a textbook.
An important limitation in employing lecture videos for learning is the challenge of quickly accessing a video segment that contains the content of interest.
Conventional video format inherently lacks non-linear navigation support like indexing and content search.
Videopoints overcomes this limitation by partitioning the video into segments discussing a sub-topic in the lecture, and presenting the segments to the users via visual index frames~\cite{tuna:developmentandevaluation, tuna-learningCompanion2017}.

The main goal of the research presented in this paper is to build a visual summary to provide a natural way to connect to a lecture video segment. These visual summaries are employed to index lecture video segments to improve navigation.  A visual summary is expected to be more informative for the user than the first frame of a video segment that is often used by default for indexing. The video summary can be used in conjunction with a text summary \cite{Shalini-2019} in a real world system like Videopoints.





The approach taken in this work is based on analyzing the video frames in a lecture video segment, identifying the most relevant images, and synthesizing a new frame that is representative of the visual objects in the video segment. In practice, we only need to analyze the {\em transition frames}, which are the video frames where the scene in the video changes based on the RGB values of corresponding pixels \cite{Hampapur95, Hanger95, Mo04, tuna:developmentandevaluation}. We outline the process of generating the visual summary. First, all images on each transition frame are identified. Text portions in a frame are identified with OCR tools and not considered to be images. Next, a "visual distance" between each pair of images is calculated based on low level image features. Finally a subset of most representative images is identified based on the image distance matrix and other factors including the size of the image and the amount of time it is displayed in the video.

 Visual summarization is illustrated in Figure {\ref{fig:teaser}}. There are four transition frames identified within a video segment. A possible visual summary is shown to the right in the figure.



\section{Background}

\begin{figure}[tbh]
  \includegraphics[width=\linewidth]{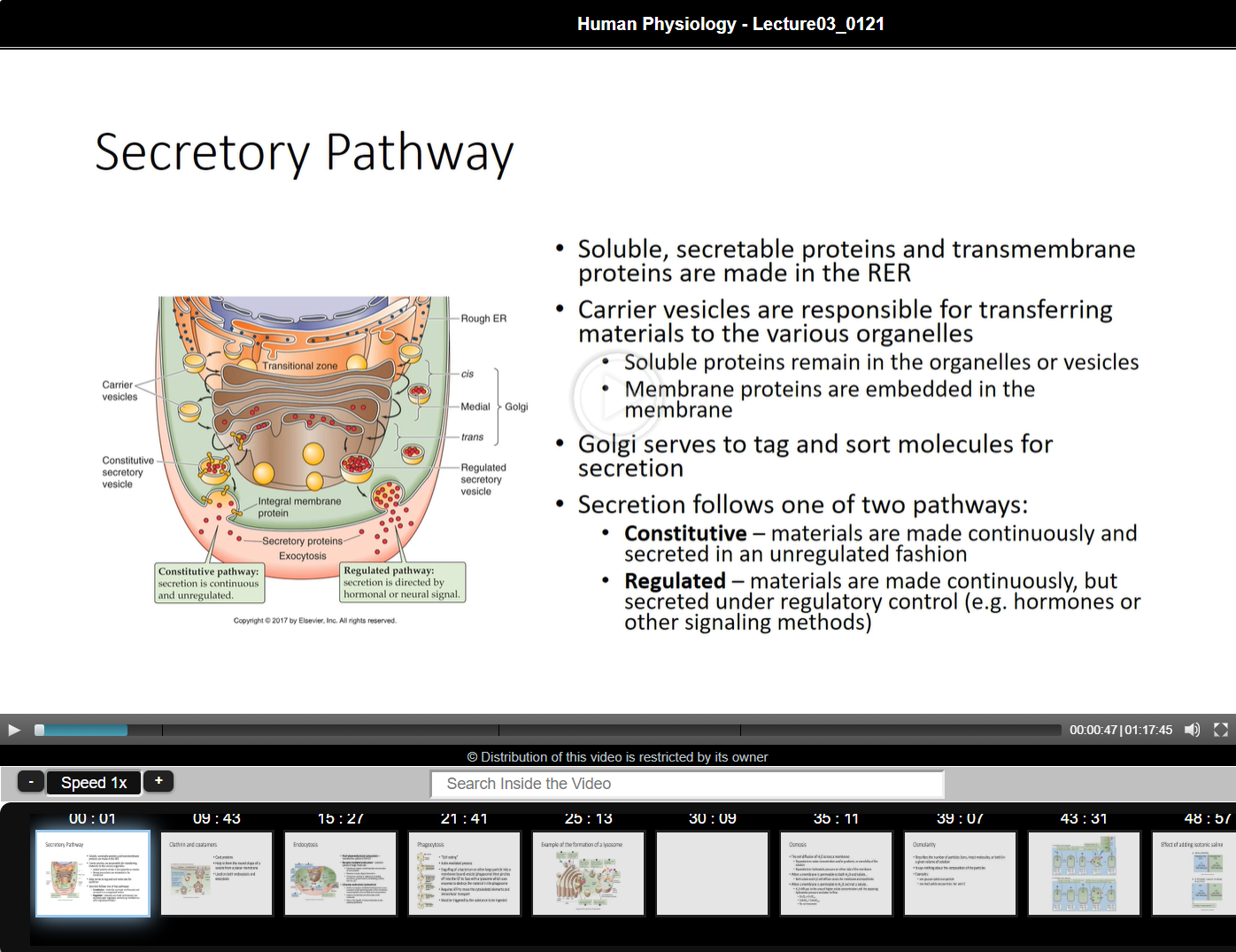}
  \caption{Videopoints Player}
  \label{fig:player}
\end{figure}

Proposed research has its roots in the Videopoints project. The project developed a lecture video framework illustrated in Figure~\ref{fig:player} that is in active use. Various aspects of this work are reported in detail in~\cite{leciaBarker:studentPerceptions,BaSu12,tuna:indexingandSearch,tuna:developmentandevaluation,tuna:topicbased,rucha:captionEditor,tuna:videobasedflipped}. We briefly describe the Videopoints player that encapsulates indexing, search, and captioning. An index panel is situated on the bottom of the player; each index point represents a new topic in the form of a screen-shot of the video at that point of time. Users can navigate different topical segments of the video by clicking on these index frames.

\begin{figure*}[bth]
\centering
\includegraphics[width=0.43\textwidth]{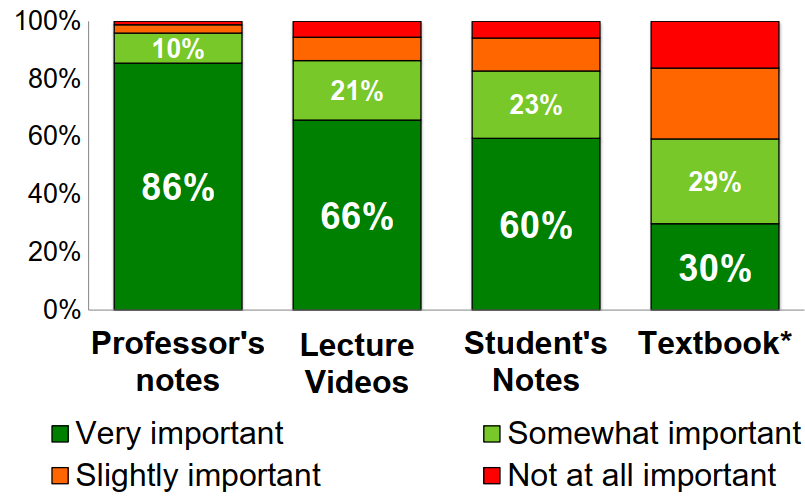}
\hspace*{0.3in}
\includegraphics[width=0.5\textwidth, height=1.9in]{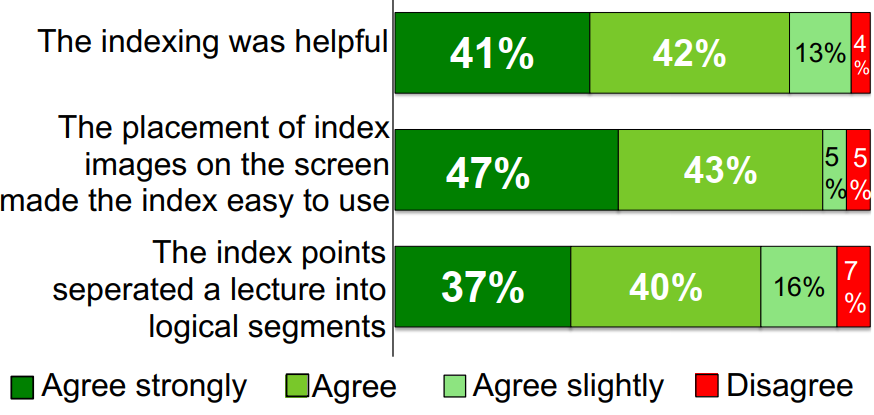}\\ (a) Student ratings of learning resources \hspace*{1in} (b)  Perception of value of indexing
\caption{Lecture video as a learning resource and its evaluation}
\label{fig:videovalue}
\end{figure*}

This video framework has been deployed at University of Houston and used for coursework across Biology, Chemistry, Computer Science, Geology, Mathematics, and Physics.  More than 3,500 students in 73 course offerings were surveyed \cite{tuna-learningCompanion2017}. Figure \ref{fig:videovalue}(a) shows that the students considered access to lecture videos and the player very important and figure \ref{fig:videovalue}(b) shows that student users were generally satisfied with the topical indexing feature.

\section{Related Work}

Video summarization is a long-studied problem in multimedia content analysis. Several projects have attempted to summarize videos to facilitate quick retrieval or browsing of large volumes of data generated by video cameras \cite{Divakaran01videosummarization,718513,1035715,Zhang1995,DeMenthon:1998:VSC:290747.290773}.  These approaches try to identify a small set of frames in a video that can convey the overall content of the video, that are then presented to the user.   These video summarization techniques have been investigated mostly for video obtained from typical video cameras. The results are not directly applicable to lecture videos or screencasts, which is the main focus of this research.

In recent years, a body of research has been developed to analyze lecture videos. Several contributions focus on finding unique transition frames or minimizing the number of frames of videos across different types of presentations like powerpoint lectures and blackboard handwriting \cite{C2,718513,C4,C5,C6}. Recent research has also focused on  methods for efficient access to educational videos with automatic identification of keywords from videos or their segments \cite{C7,Shalini-2019}.


One contribution indirectly related to our work is ViZig  \cite{C8}.  In this work visual content is extracted, classified, and presented with a direct link to the player timeline where it appeared in the video to help non-linear navigation. However, this work does not address visual summarization.

In the context of identifying images relevant for visual summarization, this research leverages existing methods for  detecting and matching interest points to establish a measure of visual similarity between images. Several interest point detection algorithms such as Harris corner detector~\cite{HARRIS88} and others such as the Hessian-Laplace, Laplacian-of-Gaussian, and the Laplacian-of-Gaussian using a Difference-of-Gaussian~\cite{LINDBERG98,52,C9,MIKO06} and numerous features such as SIFT~\cite{C9}, SURF~\cite{SURF06}, GLOH~\cite{56,57}, DAISY~\cite{DAISY08}, and their variants~\cite{54,56,57} have been proposed and their usage demonstrated to a range of image analysis applications including image matching problems. 

\section{Visual Content Summarization}
\label{sec:summarization}
\begin{figure*}[ht]
  \includegraphics[width=\textwidth]{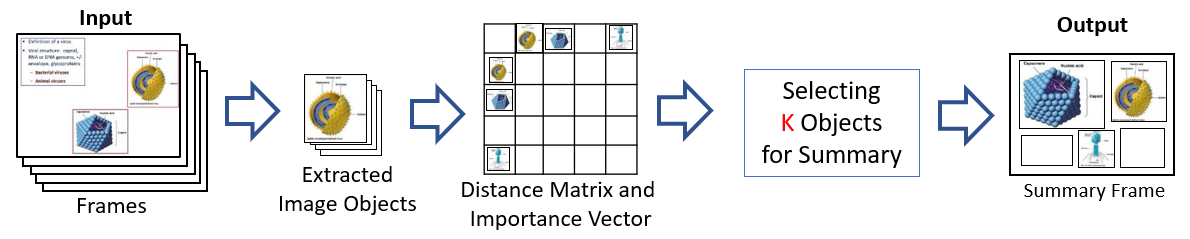}
  \caption{Algorithm Steps}
  \label{fig:algo-steps}
\end{figure*}


The main technical objective of this paper is to generate a visual summary of a lecture video or screencast, where the content on the screen consists of text and images, that change often but not continuously. Specifically, the goal is to identify a subset of the images from the frames of the video segment that best represents the content of the lecture segment. The approach taken for generating the visual summary of a lecture video segment in this paper consists of the following steps:
\begin{enumerate}
    \item Extract all images from the frames in the video segment, along with the time for which they were displayed in the video timeline.
    \item Compute the "importance" of each individual image for inclusion in the visual summary.
    \item Compute the "distance matrix" between all pairs of images based on (dis)similarity between the images.
    \item Select the subset of images that best represent the segment based on similarity and importance.
    
\end{enumerate}
These steps are illustrated in Figure \ref{fig:algo-steps}. We describe the steps 1-3 in this section which generate an importance vector, and an image-image distance matrix. The selection of representative images from a distance matrix is detailed in section {\ref{sec:selection-rep-image}}. The final selected images are placed in a single frame on uniformly sized cells. A more visually appealing arrangement is possible but beyond the scope of this work. An example set of lecture video frames and the corresponding visual summary is illustrated in Figure \ref{fig:teaser}.

\subsection{Extracting Images from a Lecture Video Segment}
\label{ssec:step1-extration}
A lecture video or screencast typically consists of a small number of unique video frames, each displayed from a few seconds to several minutes. The content of the video frame is often a {\em Powerpoint} viewgraph, but it can be an image from the web or elsewhere. Following are the steps to extract the image objects in the frames of the segment.
\begin{enumerate}
    \item {\em Identify transition frames:} The first step in this analysis is to identify {\em transition frames} where the scene in the video changes significantly based on the RGB values of the corresponding pixels. 
    Figure \ref{fig:teaser} shows the 4 transition frames in an example lecture video segment.
    \item {\em Remove text regions:} The next step is to remove text regions from the transition frames. This is to prevent blocks of text being identified as images. We employ an OCR engine for this purpose. OCR extracted text regions are grouped together in a block of text, if they are close to each other. 
    \item {\em Identify image regions:} At this stage, the transition frames consist of images and empty space. We employ a simple technique to identify image regions that proceeds as follows. A transition frame is scanned for pixel changes with a sliding window protocol. Bounding boxes enclosing image objects are identified as regions surrounded by a border with no visual content. 
\end{enumerate}

\begin{figure}
  \includegraphics[width=\linewidth]{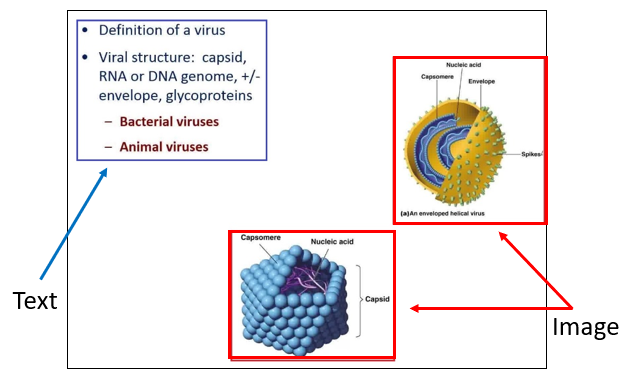}
  \caption{Extracting Images from a Lecture Video Segment}
  \label{fig:algo-step-1}
\end{figure}

A transition frame with boxes containing image objects and a text region are illustrated in
    Figure \ref{fig:algo-step-1}.
 This work leverages the framework developed for identifying transition frames and extracting text with OCR for content based indexing \cite{tuna:indexingandSearch, tuna:topicbased}. 

\subsection{Image Importance}
\label{ssec:step2-importance}
Desirability of an image to be included in a visual summary is an independent consideration from similarity to other images. This {\em importance} of an image depends on a number of factors including  the size of image, information density in the image, and the duration for which the image is visible in the lecture video. The information density is captured by number of keypoints in unit area. We have developed a heuristic definition of importance as follows. We normalize these features for all images in the segment and then calculate the importance using following equation:
\[Importance= Size * InfoDensity * Duration\] 



\subsection{Image Distance Matrix}
\label{ssec:step3-distancematrix}
An important consideration in deciding whether an image should be included in a visual summary is to quantify how similar (or different) it is to other images. In this work we calculate the similarity between each pair of extracted image objects and create a distance matrix, where
\[Distance = 1 - Similarity\]

Computing a measure of similarity between two images is a well studied problem. There are many metrics and algorithms to measure image similarity. Global measures include holistic image properties such as color and texture computed from the entire image or parts of an image, often represented as histograms. In contrast, local measures rely on identifying parts or points within an image that are unique to an image or objects within an image, which in turn can be used in computing similarity between images.   Finding similarity of visual objects in a lecture video is a unique problem as the images are typically synthetically created and do not represent any real world objects. Images often contain illustrations like diagrams, cartoons, charts, and graphs. The images often have a specific meaning in a particular domain only. Images tend to have geometrical shapes. Pairs of images, where one image is a rotated, scaled, or cropped version of another, are common in a lecture video segment. It is also common that one image is simply another image with additional visual content. 

Based on above considerations and our practical experience, we chose to focus on local features that are invariant to geometric transformations. We chose to use SIFT \cite{C9} to extract local interest points (keypoints) and the corresponding feature descriptors from an image to be used for matching interest points between images.  To compute a measure of similarity between two images, we measure:
\begin{itemize}
    \item {\em KeypointsScore.} The percentage of unique keypoints that match between the pair of images; and
    \item {\em TransformationScore.} The degree to which one image is a geometric transformation of the other.
\end{itemize}

The percentage of keypoints matched provides an indication of local similarities between two images.  We further compute an affine transformation based on matched keypoints.  The second image is transformed and aligned with the first image using the computed transformation and a pixel-wise normalized difference is measured to provide an indication of the global similarity between two images.       
The final similarity score is simply the average of these two measures, given as: 
\[Similarity= avg(KeypointsScore, Transformation Score)\] 

\section{Selection of Representative Images}
\label{sec:selection-rep-image}
 In section ~\ref{sec:summarization} we described the process of obtaining the set of images in a lecture video segment, an importance value for each image, and a distance matrix that captures the dissimlarity between pairs of images.  In this section we discuss the mechanism of identifying a small subset of those images that best represent the content of the lecture video segment. We first present a theoretical formulation of the problem and then present practical algorithms to solve it. 

\subsection{Problem Description}
Suppose a lecture video segment has $n$ images (or visual object), $V_1, V_2, V_3,$ $V_4, ..., V_n$. The goal is to construct a visual summary consisting of $m$ representative images, $R_1, R_2, R_3, R_4,...R_m$ where $m<n$.  An $nxn$ $Distance$ matrix is available where $Distance_{ij}$ captures the visual difference between images $V_i$ and $V_j$. A vector $Importance$ of size $n$ is provided where $Importance_i$ captures the importance of the corresponding image $V_i$. 

For identifying representative images for the summary, we apply two considerations: i) minimize the distance between each image not in the summary to the closest representative image in the summary, and ii) prioritize images that have more importance. Quantitatively our optimality criterion is to identify a set of representative images for which the maximum of $Distance_{ir}*Importance_i$ over all images is minimized, where $Distance_{ir}$ is the distance between image $V_i$ and the image $V_r$ in the summary that is closest to $V_i$.

\subsection{Optimal representative image selection}
We informally show that the problem of optimally selecting representative images as stated in this section is NP-hard.

Consider a highly simplified version of the problem where i)  all images are of equal importance and ii) the visual distance between any pair of images is 0 (identical) or infinite (no similarity). We can think of this version of the problem as a graph problem where each of the images $V_1, V_2, V_3,...V_n$ is a graph node. There is an edge between the nodes if the corresponding visual distance between the images is 0, and no edge if the corresponding distance is infinite. In this scenario, the problem of finding an optimal summary of $m$ representative images translates to finding $m$ graph nodes, such that all other graph nodes are adjacent to one of the nodes in this set. This is the exact formulation of the problem of finding a {\em dominating set} of a graph which is known to be NP-hard \cite{garey1979computers}. Hence the problem of identifying representative images as detailed in this section is NP-hard.

\subsection{Graph based algorithm}
While the general problem of finding the optimal visual summary is NP-hard, the real world instance of interest to us is not as daunting as the number of images in a segment is typically not that large. In most instances, even an exponential algorithm that examines all combinations of images as candidates for the visual summary can be run within a few minutes on a desktop. However, this is not a practical approach above a certain size and visual complexity of a lecture video segment.

We introduce an efficient heuristic algorithm for identifying a good set of representative images that proceeds as follows. Initially the visual summary consists of all  images in the segment. In the following step, the cost of removing each image from the summary is computed.

The cost $cost_k$ of removing image $V_k$ is computed as follows:

$$cost_k =  I_k * D_{k,p} $$

where $V_p$ is the node in the summary that has the least distance (or is most similar to) $V_k$, $I_k$ is the importance of image $V_k$ and $D_{k,p}$ is the distance between $V_k$ and $V_p$. The node with the lowest cost is removed. This step is repeated until the desired $m$ nodes are left in the summary. The complexity of this heuristic algorithm is $O(n^3)$ for a segment with $n$ images\footnote{Strictly speaking, the complexity of the algorithm as stated for illustration in the paper would be $O(n^4)$ but a trivial change that records the closest node in summary set for all other nodes would make it $O(n^3)$.}.

For the data set employed for experiments in this paper, we verified with a brute force approach that the results from the heuristic algorithm nearly always matched the optimal results.


\begin{algorithm}
\SetAlgoLined
\DontPrintSemicolon
\SetKwInOut{Input}{Input}
\Input{$V$ (Set of Images), $m$ (Number of Images in Summary), $D$ (Distance Matrix), $I$ (Importance Vector) }{}
\SetKwInOut{Output}{Output}
\Output{$S$ (Set of Summary Images) }
 $S \gets V$ \;
 \While{  $|S|$  > $m$ }{
   $costmin  \gets 1$\; \tcp*{ set to maximum possible value of $cost$}
   \ForEach {$s_j \epsilon S$}{
     $cost_j \gets FindMinimumCost(V, S, s_j,D,I)$ \;
     \If{ $cost_j$ < $costmin$ }{
       $s_{min} \gets s_j$\;
       $costmin \gets cost_j$\;
       
     }
   }
   $S \gets \{S-s_{min} \}$\;

 }
  \;

  \tcc{$FindMinimumCost$ calculates minimum cost for removing $s_j$ from Summary $S$}
  \SetKwFunction{FMC}{$FindMinimumCost$}
  \SetKwProg{Fn}{Function}{:}{}
  \Fn{\FMC{$V$, $S$, $s_j$, $D$, $I$}}{
          $S' \gets S - s_j $ \;
          $M \gets V - S'$ \;
         $MinimumCost \gets 1$\; \tcp*{ set to maximum possible value of $cost$}
         
         \ForEach {$m \epsilon M$}{
            $cost_{m} \gets 1$\;
            \ForEach {$s \epsilon $S'}{
                \If{$cost_{m} > I_{m} * D_{s,m}$}{
                    $cost_{m} \gets I_{m} * D_{s,m}$
                }
                \If{$cost_{m}<MinimumCost$}{
                    $MinimumCost \gets cost_{m}$
                }
            }
         }
        \KwRet{$MinimumCost$} \;
  }

\caption{Graph Based Algorithm}
\end{algorithm}

\section{EVALUATION}
\label{sec:evaluation}
The results of visual summarization were tested and evaluated in the context of Videopoints portal for lecture videos that is widely used at the University of Houston. Results are presented for the graph based algorithm presented in this paper, as well as the K-medoid clustering based algorithm. We describe the process of ground truth collection and the metrics used to evaluate the quality of the summaries generated by our algorithms.
 
\subsection{Ground truth collection}
 Ground truth was collected with a survey of users of the Videopoints system. 

\subsubsection*{Dataset}
\label{ssec:dataset}
 We selected 40 segments from 4 subjects taught at the University of Houston. The subject areas, in decreasing order of the number  of selected  segments, are Biology, Geoscience, Computer Science and Chemistry. The segments were approximately 15 minutes long on average, and contained approximately 12 images on average. Segments with 5 or fewer images were not selected for evaluation. 
 For our experiments, the algorithms were configured to provide a visual summary of up to 4 images, and the task of selecting a 4 image summary is trivial or relatively easy in these cases. More details on the data set are listed in Table~\ref{tab:video-stats}.


%

\begin{table}[ht]
\caption{Statistics on video segments for evaluation}
\label{tab:video-stats}

\begin{tabular}{|  l | c| c| }
\hline
\textbf{} & \textbf{Duration (mins)} & \textbf{Total Images} \\ \hline
Min & 2.87 & 6 \\ \hline
Max & 36.03  & 30 \\ \hline
Average & 14.99 & 12.28 \\ \hline
Median & 14.04 & 11 \\ \hline
\end{tabular}%
\end{table}

 \begin{figure*}[htb]
  \includegraphics[width=.8\textwidth]{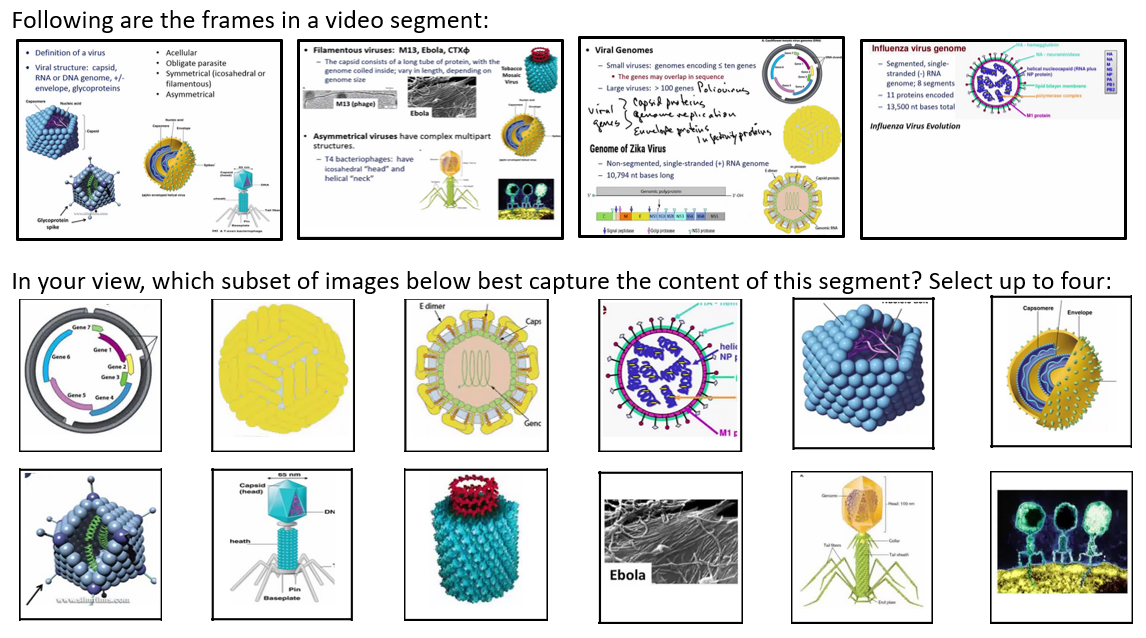}
  \caption{Interface to identify images for the ground truth visual summary}
  \label{fig:survey1}
\end{figure*}

\subsubsection*{Survey}
\label{ssec:survey}
\begin{figure}
  \includegraphics[width=.8\linewidth]{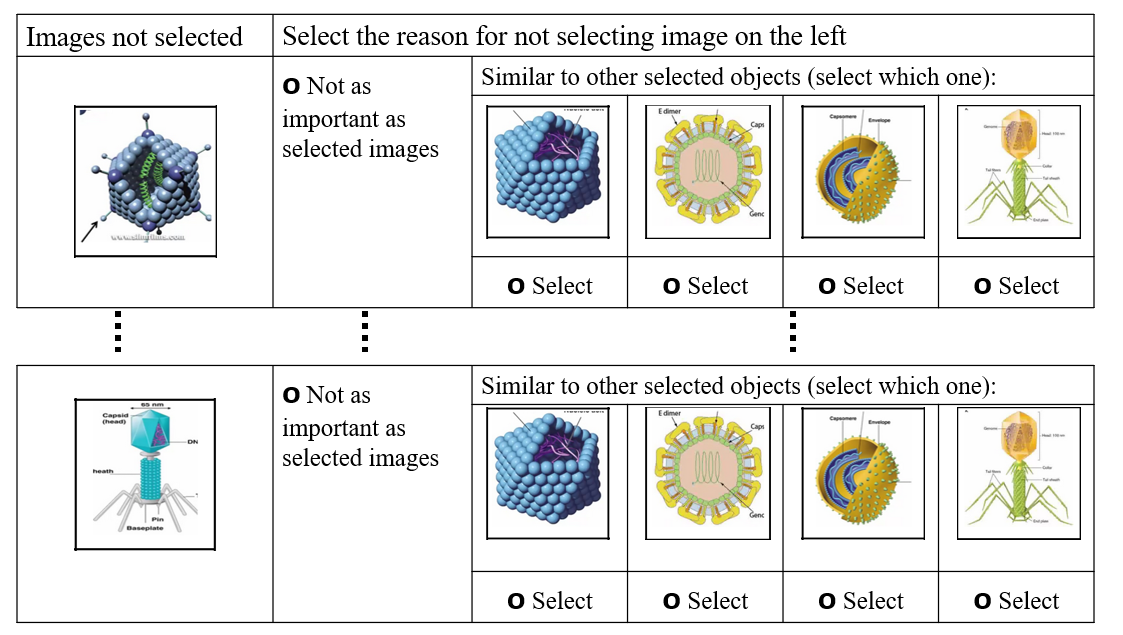}
  \caption{Interface to identify the reasons for not selecting specific images for the visual summary (truncated to show only 4 images) }
  \label{fig:survey2}
 \end{figure}

\begin{figure}
  \includegraphics[width=\linewidth]{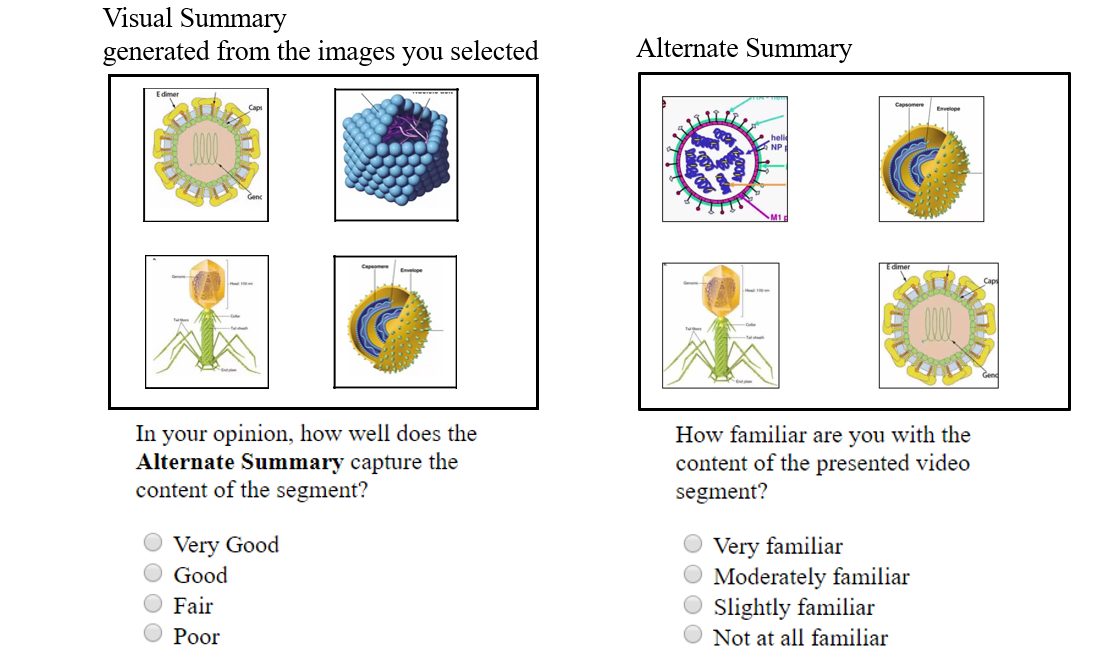}
  \caption{Interface to rate the quality of an automatically generated summary and to rate user familiarity with the subject area}
  \label{fig:survey3}
\end{figure}

 We developed a web-based survey tool and process to collect ground truth that proceeds as follows:
 \begin{enumerate}
     \item 
 A survey participant is first shown all distinct images extracted from a video segment, and asked to select a set of 4 images that best represent the segment. This is illustrated with an example in Figure~\ref{fig:survey1}.
 
 \item
Next, for every image a participant did {\em not} select for the  visual summary, they are asked if the primary reason was i) it was not as important as selected images, or ii) it was similar to one of the selected images. An example is shown in Figure \ref{fig:survey2}.
\item
Finally the participant is presented with the images automatically selected by our algorithm and asked to judge the quality of this summary on a 4 points scale with 1 as ''Very Good'' and 4 as ''Poor''. In the same step, they are asked to rate their familiarity with the content of the lecture video on a 4 point scale, with 1 as ''Very Familiar'' and 4 as ''Not at all Familiar''. An example  is shown in Figure~\ref{fig:survey3}.
  \end{enumerate}

\subsubsection*{Participants}
\label{ssec:participants}
Students and instructors were invited to participate in the survey. Since domain knowledge in the subject area is an important factor, we focused on recruiting instructors and students in the subject areas of the survey. The participants self reported areas that they were familiar with, and were assigned  segments in those areas.
The survey had 30 participants. Each segment was ranked by 5.75 participants on average, with 6 being the maximum and 3 being the minimum for any segment. 

\subsection{Evaluation Methodology}
\label{ssec:groundtruth}
Evaluation of a visual summarization algorithm is challenging for a number of reasons:
\begin{enumerate}
    \item Multiple images can express the same concept and hence a user may consider two or more images equally representative of a subtopic in a video segment. To capture this aspect, we allow users to specify such perceived similarity in the ground truth collection survey.
    
    \item There may be no consensus on ground truth. Users often differ significantly on their assessment of the best set of images that summarize a video segment.
   
    \item Different sets of images may represent the content of a video segment equally well, even if they have few images, if any, in common.
\end{enumerate}

We have developed an evaluation methodology to address these factors. We outline the salient aspects of our approach to evaluation:



\subsubsection*{Individual images selected by participants:}
During ground truth collection, different survey participants pick different images for a visual summary. We perform evaluation on a subset of images that are:
\begin{itemize}

    \item Selected most often by participants. In the experiments presented in this paper, the set of 4 most selected images was used. The maximum size of visual summary generated by our algorithms was also set to 4 images. 
    
    \item Selected by any participant. 
    
\end{itemize}

\subsubsection*{Grouping similar images:}
In the ground truth collection process image similarity information was collected in the following way. A participant was allowed to specify that they did not pick a specific image, say X, for the summary, because it was similar to another image, say Y, that they did include in the summary. We use this information to group similar images. Result are presented that take this similarity into account. For example in the above scenario, if ground truth contains Y, then the algorithm is scored identically if it selects X or Y as part of the visual summary.

\subsubsection*{Custom Metrics:}
In addition to presenting results for standard metrics (accuracy, precision, recall, F1-measure), we analyze the data to answer the following questions:
\begin{itemize}
    \item What percentage of the summary images selected by the algorithm were not selected by any participant and not considered similar to any image selected by a participant?
    \item For what percentage of the images selected by 3 or more survey participants, the algorithm did not select the image or an image considered similar to it in the summary?
\end{itemize}

These question provide additional insight in the automated summary selection process given the fuzzy nature of the ground truth.

\subsubsection*{Perception of automated summaries:} Finally, we present results on the perceived quality of the algorithm generated summary. This is an important measure as it is not uncommon to have a high quality visual summary that is significantly different from a summary provided by a human participant.

\section{Results} 
This work has developed a framework with a graph based algorithm to automatically select a set of images as a visual summary for a lecture video segment. For the purpose of evaluation, the framework was configured to select up to 4 images for the summary.
 
\subsection{Traditional Performance Metrics}

We evaluate our algorithm to automatically select a set of images with 4 different formulations
\begin{description}
\item[Top-4 Selected:] The ground truth is composed of 4 images that were selected most often by the survey participants.
\item[All Selected:] The ground truth is composed of union of all images selected by the survey participants.
\item[Top-4 Selected with Grouping:] The images are first divided into groups based on any user specifying that one image is similar to another, as discussed in Section \ref{sec:evaluation}. The ground truth is composed of 4 groups whose members were selected most often by the survey participants. 
\item[All Selected with Grouping:] The images are first divided into groups based on a user specifying that one image is similar to another, as discussed in Section \ref{sec:evaluation}. The ground truth is composed of all groups for which any image was selected by any survey participant.
\end{description}

\begin{figure}[ht]
  \includegraphics[width=\linewidth]{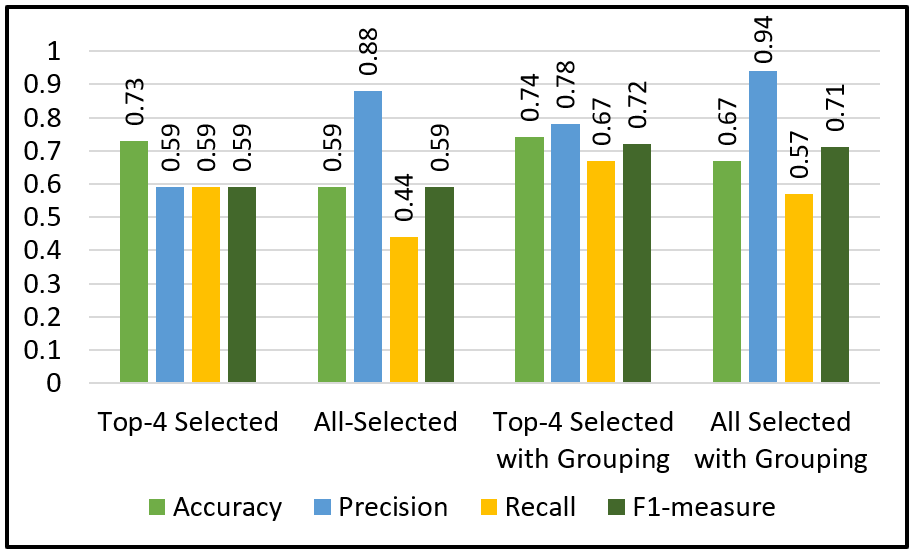}
  \caption{Performance of graph based visual summarization algorithm}
  \label{fig:std-metric}
 \end{figure}

 Figure \ref{fig:std-metric} shows algorithm performance with the formulations stated above. We make a few observations: i) The scores are significantly higher with grouping indicating the role played by user selected similar images, ii) precision is significantly higher but recall is lower when all participant choices are added to the ground truth, and  iii) precision reaches a high 0.94 for "All Selected with Grouping" implying that most algorithm choices found some agreement with at least one survey participant.

 \subsection{Comparison with Clustering}
\begin{figure}[ht]
  \includegraphics[width=\linewidth]{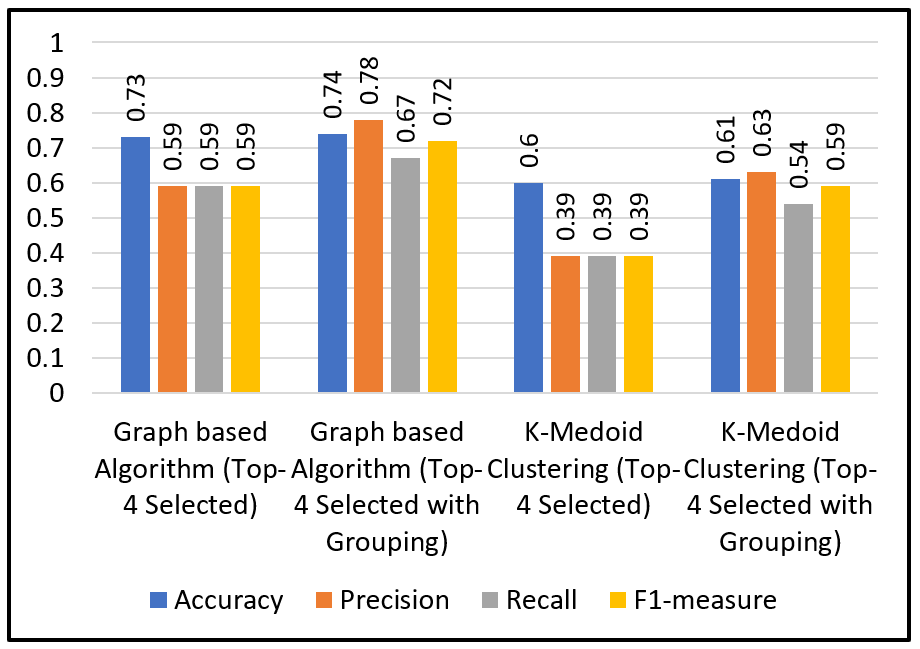}
  \caption{Performance of graph based algorithm vs K-medoid clustering}
  \label{fig:std-metric-3}
 \end{figure}

We compare the results from our graph based algorithm with K-medoid clustering algorithm \cite{Madhulatha-2011}. We consider K-medoid to be a reasonable clustering algorithm in this scenario as it has been used in the context of  image retrieval and face recognition \cite{Sarthak-2014,Yushi-2010}. The results are presented in Figure \ref{fig:std-metric-3}. we note that our graph based algorithm yields slightly to significantly better performance on all metrics in the two scenarios that were compared.


\subsection{Custom Metrics}
We now address two specific questions that provide further insight into the ability of algorithms to correctly pick summary images.

\begin{itemize}
    \item 12.5\% of the summary images selected by the algorithm were not selected by any participant. Of these, only 1.88\%  were not selected by any participant and not considered similar to any image selected by a participant.
    \item  40\% of the images selected by 3 or more participants were not selected by the algorithm. Around half of these, i.e., 20\%,  were considered similar to an image selected by the algorithm.
    
\end{itemize}
In brief, the algorithm selected images were almost always considered relevant by at least one human, but top rated human images were not selected by the algorithm in several cases.



\subsection{User Perceptions}
\begin{figure}[ht]
  \includegraphics[width=\linewidth]{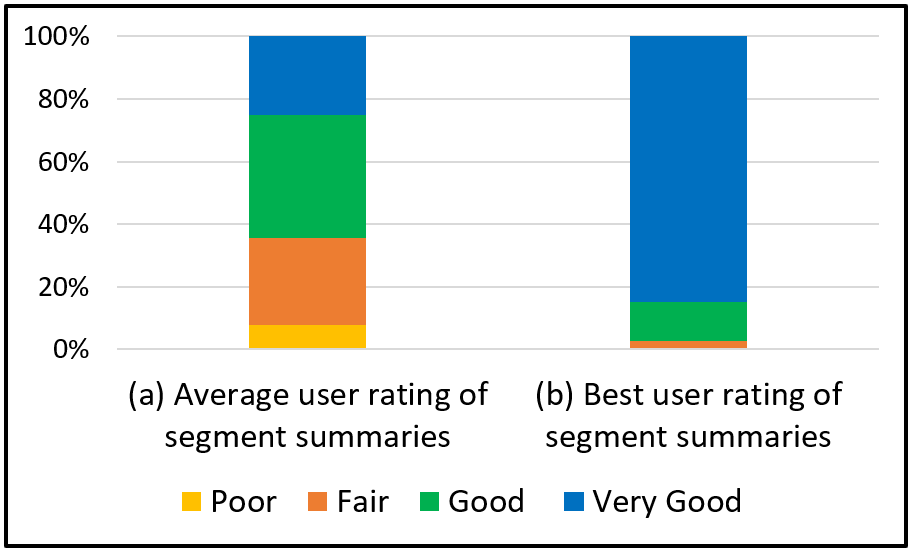}
  \caption{User perception of quality of algorithm generated summary}
  \label{fig:feedback-quality-new}
 \end{figure}

User survey participants were provided with the algorithm  selected visual summary and asked to rank its quality on a 4 point scale from "Poor" to "Very Good".  Figure \ref{fig:feedback-quality-new} plots the results.  Figure \ref{fig:feedback-quality-new} (a) shows that the users considered around 65\% of segment summaries to be "Good" or "Very Good", and around 7.6\% of summaries to be "Poor".  Figure \ref{fig:feedback-quality-new} (b) plots the most positive review for a summary among all users who rated it. We observe that 85\% of the algorithm generated summaries were rated to be "Very Good" by at least one user, and virtually all summaries were rated "Good" or "Very Good" by at least one user. Since evaluation of a summary is qualitative, it is an important result that at least one user was largely satisfied with an automatically generated summary for most segments. 

We also surveyed the users for their level of familiarity with the content of the lecture videos they rated. Around 54.5\% of the users considered themselves to be "Very Familiar", 25.2\% were "Familiar", 12.6\% were "Slightly Familiar" and 8.1\% were "Not at all Familiar". We observe that our efforts to match users with content matter they were familiar with were mostly successful. We verified that the small number of users with low familiarity do not meaningfully impact our results.

\section{Conclusions and Future Work}

This paper presents a novel approach to use low-level image features to create a summary of visual contents extracted from a lecture video segment. The results are encouraging based on quantitative metrics as well as user perception of the quality of visual summaries. At the same time, there is also significant room for improvement.

Ongoing work is identifying and classifying the underlying causes of errors in visual summaries. Future work will focus on improving the accuracy and relevance of extracted summaries. Research directions under consideration include i) alternate image similarity measures, ii) analysis of high level semantic features, and iii) integrated text and image analysis. We also plan to substantially enhance the ground truth with additional surveys to be able to make more accurate assessments.


\begin{acks}
The bulk of the Videopoints framework that we employed for evaluation was developed by a team led by Dr. Tayfun Tuna. Dr Thamar Solorio and Dr Ioannis Konstantinidis made numerous helpful suggestions on this project.
\end{acks}

\bibliographystyle{ACM-Reference-Format}
\bibliography{acmart}


 

\end{document}